\documentclass[runningheads]{Format/llncs}
\usepackage[only-used=true, single=false]{acro}
\usepackage[ruled,vlined,linesnumbered]{algorithm2e}
\usepackage{amsfonts}
\usepackage{amsmath, amssymb}
\usepackage{array}
\usepackage{bm}
\usepackage{color}
\usepackage[utf8]{inputenc} % required for especial symbols in the text
\usepackage[T1]{fontenc}    % required for the accent in author
\usepackage{graphicx}
\usepackage{subfigure}
\usepackage{url}
\usepackage{mathtools}

\newcommand*{\cref}[1]{Chapter~\ref{#1}}
\newcommand*{\eref}[1]{(\ref{#1})}

\newcommand*{\fref}[1]{Figure~\ref{#1}}

\newcommand{\real}{\mathbb{R}}
\newcommand{\sphere}{\mathbb{S}}
\newcommand{\rotmat}{\mathrm{SO}}

\newcommand{\workspace}{\mathcal{W}}
\newcommand{\object}{\mathcal{O}}

\newcommand{\graspmatrix}{\mathbf{G}}

\newcommand{\vy}{\mathbf{y}}

\newcommand{\pos}{x}
\newcommand{\vel}{\dot{x}}

\newcommand{\vq}{\mathbf{q}}
\newcommand{\vdq}{\mathbf{\dot{q}}}

\newcommand{\vgoal}{\mathbf{g}}
\newcommand{\Askill}{\mathcal{S}_a}
\newcommand{\Rskill}{\mathcal{S}_r}

\newcommand{\adaptation}[1]{\mathbf{f}{#1}(\boldsymbol{\cdot})}

\DeclareAcronym{1D}{
  short = 1D,
  long  = one-dimensional
}
\DeclareAcronym{2D}{
  short = 2D,
  long  = two-dimensional
}
\DeclareAcronym{3D}{
  short = 3D,
  long  = three-dimensional
}
\DeclareAcronym{AI}{
  short = AI,
  long  = artificial intelligence
}
\DeclareAcronym{CAN}{
  short = CAN,
  long  = controller area network
}
\DeclareAcronym{CF}{
  short = CF,
  long  = coupling force
}
\DeclareAcronym{CMP}{
  short = CMP,
  long  = compliant movement primitive
}
\DeclareAcronym{DMP}{
  short = DMP,
  long  = dynamic movement primitive
}
\DeclareAcronym{DoF}{
  short = DoF,
  long  = degree of freedom,
  long-plural-form = degrees of freedom
}
\DeclareAcronym{DS}{
  short = DS,
  long  = dynamical system
}
\DeclareAcronym{ECR}{
  short = ECR,
  long  = Edinburgh Centre for Robotics
}
\DeclareAcronym{EM}{
  short = EM,
  long  = expectation-maximisation
}
\DeclareAcronym{FK}{
  short = FK,
  long  = forwad kinematics
}
\DeclareAcronym{GMM}{
  short = GMM,
  long  = Gaussian mixture model
}
\DeclareAcronym{GMR}{
  short = GMR,
  long  = Gaussian mixture regression
}
\DeclareAcronym{GPR}{
  short = GPR,
  long  = Gaussian process regression
}
\DeclareAcronym{HMM}{
  short = HMM,
  long  = hidden Markov model
}
\DeclareAcronym{HRI}{
  short = HRI,
  long  = human-robot interaction
}
\DeclareAcronym{HSMM}{
  short = HSMM,
  long  = hidden semi-Markov model
}
\DeclareAcronym{KL}{
  short = KL,
  long  = Kullback-Leibler
}
\DeclareAcronym{IGMM}{
  short = IGMM,
  long  = infinite Gaussian mixture model
}
\DeclareAcronym{IIT}{
  short = IIT,
  long  = Italian Institute of Technology
}
\DeclareAcronym{IK}{
  short = IK,
  long  = inverse kinematics
}
\DeclareAcronym{ILC}{
  short = ILC,
  long  = iterative learning control
}
\DeclareAcronym{IR}{
  short = IR,
  long  = infra-red
}
\DeclareAcronym{LbD}{
  short = LbD,
  long  = learning by demonstration
}
\DeclareAcronym{LED}{
  short = LED,
  long  = light-emitting diode
}
\DeclareAcronym{LMS}{
  short = LMS,
  long  = least mean squares
}
\DeclareAcronym{LS}{
  short = LS,
  long  = linear square
}
\DeclareAcronym{LWPR}{
  short = LWPR,
  long  = locally weighted projection regression
}
\DeclareAcronym{LWR}{
  short = LWR,
  long  = locally weighted regression
}
\DeclareAcronym{PD}{
  short = PD,
  long  = proportional-derivative
}
\DeclareAcronym{RBF}{
  short = RBF,
  long  = radial basis function
}
\DeclareAcronym{RFWR}{
  short = RFWR,
  long  = receptive field weighted regression
}
\DeclareAcronym{RL}{
  short = RL,
  long  = reinforcement learning
}
\DeclareAcronym{ROS}{
  short = ROS,
  long  = robot operating system
}
\DeclareAcronym{WP}{
  short = WP,
  long  = work package
}
\DeclareAcronym{YARP}{
  short = YARP,
  long  = yet another robotic platform
}

\begin{document}
  \title{Learning and Composing Primitive Skills for Dual-arm Manipulation}
  \author{\`Eric Pairet\inst{1,2}\orcidID{0000-0002-3363-0426} \and
  Paola Ard\'on\inst{1,2}\orcidID{0000-0002-3026-0706} \and \\
  Michael Mistry\inst{1}\orcidID{0000-0003-3979-922X} \and
  Yvan Petillot\inst{2}\orcidID{0000-0002-1596-289X}}

  \authorrunning{{\`E}. Pairet et al.}
  \institute{Institute of Perception, Action and Behaviour, University of Edinburgh, UK  \email{eric.pairet@ed.ac.uk} \email{paola.ardon@ed.ac.uk} \email{mmistry@inf.ed.ac.uk} \and
  Engineering and Physical Sciences, Heriot-Watt University, UK \email{y.r.petillot@hw.ac.uk}}
  \maketitle              % typeset the header of the contribution
	\begin{abstract}
    In an attempt to confer robots with complex manipulation capabilities, dual-arm anthropomorphic systems have become an important research topic in the robotics community. Most approaches in the literature rely upon a great understanding of the dynamics underlying the system's behaviour and yet offer limited autonomous generalisation capabilities. To address these limitations, this work proposes a modelisation for dual-arm manipulators based on \aclp{DMP} laying in two orthogonal spaces. The modularity and learning capabilities of this model are leveraged to formulate a novel end-to-end learning-based framework which (i)~learns a library of primitive skills from human demonstrations, and (ii)~composes such knowledge simultaneously and sequentially to confront novel scenarios. The feasibility of the proposal is evaluated by teaching the iCub humanoid the basic skills to succeed on simulated dual-arm pick-and-place tasks. The results suggest the learning and generalisation capabilities of the proposed framework extend to autonomously conduct undemonstrated dual-arm manipulation tasks.
\end{abstract}

	\section{INTRODUCTION} \label{sec:introduction}
    Complex manipulation tasks can be achieved by endowing anthropomorphic robots with dual-arm manipulation capabilities. 
    Bi-manual arrangements extend the systems competences to efficiently perform tasks involving large objects or assembling multi-component elements without external assistance. These systems not only deal with the challenges of single-arm manipulators, such as trajectory planning and environmental interaction, but also require an accurate synchronisation between arms to avoid breaking or exposing the handled object to stress.

    Traditional approaches have addressed the aforementioned challenges by means of control and planning-based methods~\cite{smith2012dual}. These methods depend upon an excellent understanding of the exact model underlying the system’s and task's dynamics, which are commonly approximated to make the calculations computationally tractable~\cite{pairet2018uncertainty}. On top of that, some of these methods lack scalability and generalisation capabilities, involving hand-defining all possible scenarios and actions~\cite{billard2008robot,argall2009survey}. All these issues have motivated the use of more natural techniques for robot programming, such as \acf{LbD}, in which a human movement is recorded to be later reproduced by a robot. %Such an approach allows non-robotics-experts to easily interact, teach and modify the robot’s behaviour.

    \begin{figure}[t!]
        \centering
        {\includegraphics[height=3.2cm]{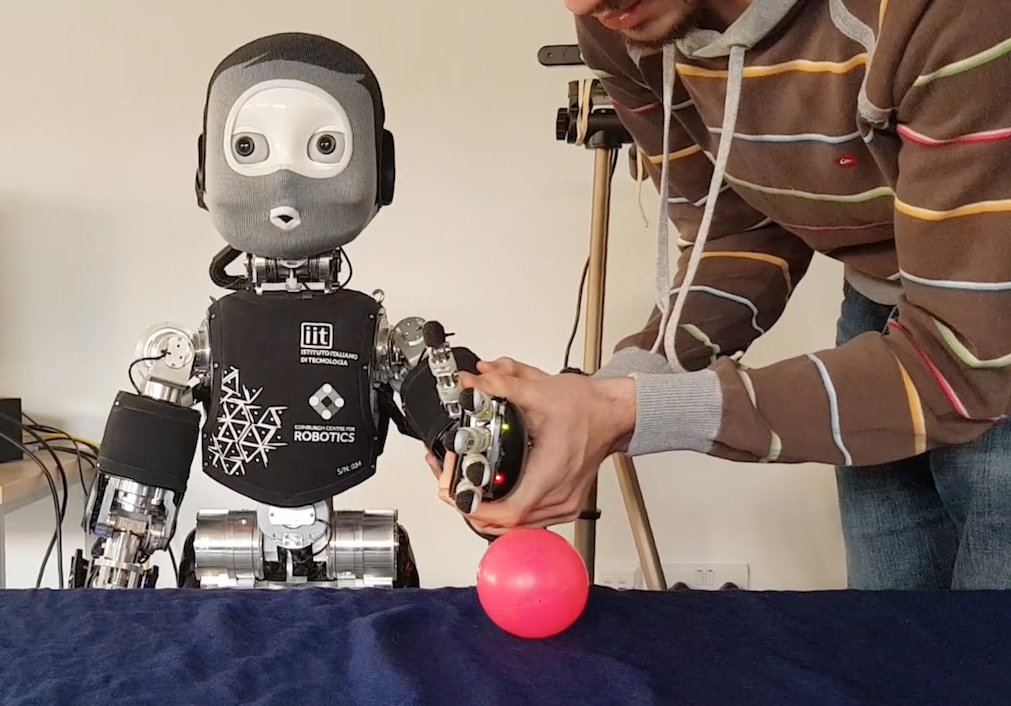}\label{fig:teaching_a}} \quad
        {\includegraphics[height=3.2cm]{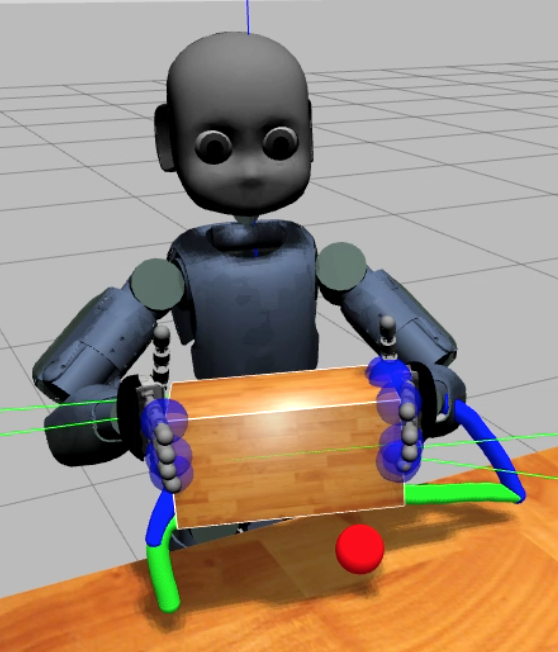}}
        \caption{iCub humanoid learning to contour an obstacle through kinaesthetic guiding (left), and composing multiple skills to conduct a dual-arm pick-and-place task (right).}
        \label{fig:lbd}
    \end{figure}

    Despite the encouraging possibilities offered by adopting human knowledge for robot control, teaching complex systems, such as dual-arm manipulators, to respond and adapt to a broad case of scenarios remains an open challenge. Particularly, it is expected from a dual-arm system to generalise the provided demonstrations to confront novel scenarios in (a)~the task space to deal with the changing requirements about trajectory planning and environmental interaction, and (b)~the relative space to ensure the essential synchronisation between arms~\cite{pairet2018learning}. However, current learning-based architectures in the literature pursuing autonomy and robustness against the dynamic and unpredictable real-world environments are limited to single-arm arrangements~\cite{pastor2009learning,rai2014learning,bajcsy2018learning}. Contrarily, learning-based frameworks for dual-arm robots do not generalise to undemonstrated states, thus being limited to highly controlled scenarios \cite{zollner2004programming,gams2014coupling,topp2017knowledge}.

    This paper presents a novel learning-based framework which endows a dual-arm system with a real-time and generalisable method for manipulation in undemonstrated environments (see \fref{fig:lbd}). The framework models a dual-arm manipulator with a set of \aclp{DMP} laying in two orthogonal spaces to tackle the task's requirements separately from the synchronisation constraints. The modularity of the \acp{DMP} is leveraged to (i)~create a library of primitive skills from human demonstrations, and (ii)~exploit primitive skills simultaneously and sequentially to create complex behaviours. The potential of the proposal is demonstrated in simulation after recording skills with the iCub humanoid through kinaesthetic guiding. The results suggest the proposal's suitability to endow a dual-arm robot with the necessary learning and generalisation capabilities to autonomously address novel manipulation tasks.

    %The remainder of this paper firstly gives an overview of the proposed framework's fundaments. It then presents the formulated modelisation for a dual-arm system and introduces the designed learning-based framework. Finally, this paper reports the conducted experimental evaluation and concludes with some potential ideas for future work.
	%\section{DUAL-ARM DYNAMIC MOVEMENT PRIMITIVES} \label{sec:extension}
\section{DUAL-ARM SYSTEM MODELISATION} \label{sec:extension}
    This paper pursues an end-to-end learning-based framework which endows a dual-arm system with enhanced generalisation capabilities, meets the synchronisation constraints, and is easily programmable by non-robotics-experts. This work addresses all these requirements by means of learnable and composable primitive skills represented as \acfp{DMP}~\cite{ijspeert2013dynamical}. This section firstly overviews \acp{DMP} and its use in the literature. It then introduces the proposed typology of actions in a dual-arm system, which allows leveraging the strengths of a \ac{DMP}-based modelisation in the dual-arm context.

    % ===============================
    % ===============================
    % ===============================
    \subsection{Dynamic Movement Primitives} \label{sec:dmps}
        \acp{DMP} are a versatile tool for modelling and learning complex motions. They describe the dynamics of a primitive skill as a spring-damper system under the effect of a virtual external force called coupling term. This coupling term allows for learning and reproducing any dynamical behaviour, i.e. primitive skill. Importantly, (a)~coupling terms can be learnt from human demonstrations, (b)~they can be efficiently learned and generated, (c)~a unique demonstration is already generalisable, (d)~convergence to the goal is guaranteed, and (e)~their representation is translation and time-invariant. Because of all these properties, \acp{DMP} are adapted to constitute the fundamental building blocks of this work. Next follows an introduction about \acp{DMP} and their usage to encode positional and orientational dynamics, and an overview of some coupling terms in the literature.
        
        % ===============================
        % ===============================
        % ===============================
        \subsubsection{Positional Dynamics.}
            Let the positional state of a one-\ac{DoF} system be defined by its position, linear velocity and acceleration. Then, the system's state transition is defined with non-linear differential equations as:
            \begin{align}
                \tau \dot{z} &= \alpha_x (\beta_x (g_{x} - \pos) - z) + f{_{{x}}}(\boldsymbol{\cdot}), \label{eq:dmp_1} \\
                \tau \vel &= z, \label{eq:dmp_2}
            \end{align}
            where $\tau$ is a scaling factor for time, $\pos$ is the system's position, $z$ and $\dot{z}$ respectively are the scaled velocity and accelaration, $\alpha_x$ and $\beta_x$ are constants defining the positional system's dynamics, $g_{x}$ is the model's attractor, and ${f{_{{x}}}(\boldsymbol{\cdot})}$ is the coupling term. The coupling term applying at multiple \acp{DoF} at once is defined as $\adaptation{_{{x}}}$. The system will converge to $g_{x}$ with critically damped dynamics and null velocity when ${\tau > 0}$, ${\alpha_x > 0}$, $\beta_x > 0$ and ${\beta_x = \alpha_x / 4}$~\cite{ijspeert2013dynamical}.
    
        % ===============================
        % ===============================
        % ===============================
        \subsubsection{Orientational Dynamics.}
            %For rotations, it is important to choose a representation with no singularities and whose differentiation is numerically stable. However, there is no minimal definition of orientation lying in $\real^3$. One alternative uses rotation matrices ${\mathbf{R} \in \rotmat(3)}$ and individually describes the numerical change in each of the nine values of $\mathbf{R}$ as presented in~\eref{eq:dmp_1}-\eref{eq:dmp_2}. This method, however, does not guarantee that the orthogonality requirements for rotation matrices are met at all times.
    
            A possible representation of orientations is the unit quaternion ${\vq \in \real^4 = \sphere^3}$~\cite{ude2014orientation}. They encode orientations of a system as a whole, thus ensuring the stability of the orientational dynamics integration. Let the current orientational state of a system be defined by its orientation, angular velocity and acceleration. Then, the orientational state transition is described by the following non-linear differential equations:
            \begin{align}
               \tau \bm{\dot{\eta}} &= \alpha_q (\beta_q \; 2 \log (\vgoal_{q} \ast \bar{\vq}) - \bm{\eta}) + \adaptation{_{{q}}}, \label{eq:q_dmp_1} \\
               \tau \vdq &= \frac{1}{2} \bm{\eta} \ast \vq, \label{eq:q_dmp_2}
            \end{align}
            where $\vq$ is the system's orientation, $\bm{\eta}$ and $\bm{\dot{\eta}}$ respectively are the scaled angular velocity and acceleration, $\alpha_q$ and $\beta_q$ are constants defining the system's dynamics, $\vgoal_{q} \in \sphere^3$ is the model's attractor, and ${\adaptation{_{{q}}} \in \real^4}$ is the coupling term. The operators $\log(\cdot)$, $\ast$, and $\bar{\vq}$ denote the logarithm, multiplication and conjugate operations for quaternions, respectively.
    
        % ===============================
        % ===============================
        % ===============================
        \subsubsection{Coupling Terms.}
            Coupling terms describe the system's behaviour, thus being useful to learn and retrieve any primitive skill. They are commonly used to encode the positional~\cite{ijspeert2013dynamical} and orientational~\cite{ude2014orientation} dynamics of a motion. Coupling terms are modelled in each dimension as a weighted linear combination of non-linear \acp{RBF} distributed along the trajectory. Thus, learning a certain movement relies on finding the weights of the \acp{RBF} which closely reproduce a demonstrated skill. 
            
            More complex behaviours may be achieved by exploiting an additional coupling term simultaneously with the motion-encoding one. This approach has been used to avoid joint limits and constraining the robot's workspace via repulsive forces pushing the system away from these limits~\cite{gams2009line}. Coupling terms have also been leveraged for obstacle avoidance with an analytic biologically-inspired approach describing how humans steer around obstacles~\cite{hoffmann2009biologically,rai2014learning}. Another use is for environmental and self-interaction purposes by means of a controller tracking a desired force profile~\cite{gams2014coupling}. To the best of the authors' knowledge, the practice of using coupling terms simultaneously has been limited to two primitive skills acting on the same frame or space~\cite{pastor2009learning}. Contrarily, this work further exploits the \ac{DMP} modularity to describe a dual-arm system in two orthogonal spaces with the purpose of facing complex scenarios by composing multiple coupling terms.

    % ===============================
    % ===============================
    % ===============================
    \subsection{Dual-arm Primitive Skills Taxonomy} \label{sec:formulation_taxonomy}
        Skills for single-arm manipulation have been well analysed in the robotics community. While some of this knowledge can be extrapolated for a dual-arm manipulator as a whole, their complexity resides in the arms interaction. In the context of manipulation via a dual-arm system, a possible classification of any primitive skill falls into two groups: (a)~absolute skills, which imply a change of configuration of the manipulated object in the Cartesian or absolute space $\Askill$, e.g. move or turn an object in a particular manner, and (b)~relative skills, which exert an action on the manipulated object in the object or relative space $\Rskill$, e.g. opening a bottle's screw cap, or hold a parcel employing force contact.

        Each type of primitive skill uniquely produces movement in its space since they lay in orthogonal spaces such that $\Askill \perp \Rskill$. It is natural to expect from a dual-arm system to simultaneously carry out, at least, one absolute and one relative skill to accomplish a task. Let us analyse the task of moving a bottle to a particular position while opening its screw cap. Both end-effectors synchronously move to reach the desired configuration (absolute skill). At the same time, the left end-effector is constrained to hold the bottle upright (relative skill), while the right end-effector unscrews the cap (relative skill).

    % ===============================
    % ===============================
    % ===============================
    \subsection{Dual-arm DMP-based Modelisation}
\begin{figure}[b!]
            \centering
            \includegraphics[width=5cm]{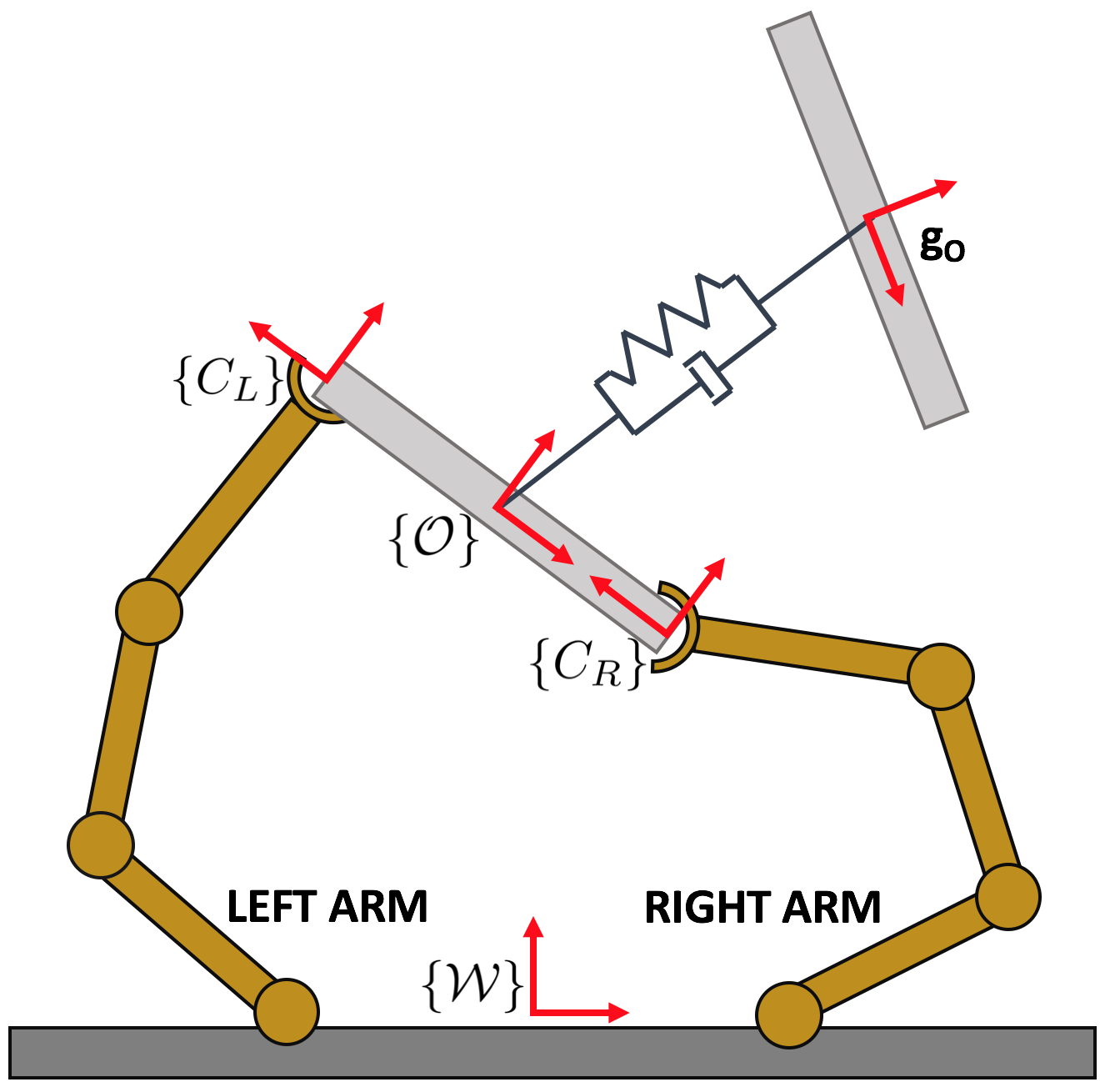}
            \caption{\ac{DMP}-based modelisation of a closed-chain dual-arm system in the absolute and relative spaces. This model is extended to deal with rotational dynamics.}
            \label{fig:system}
        \end{figure}

        Given the variety of primitive skills that a dual-arm system can execute, this work seeks to model the robotic platform in a generalisable yet modular fashion, which accounts for both absolute and relative skills. To this aim, let us consider the closed kinematic chain depicted in \fref{fig:system} operating in a \ac{3D} workspace ${\workspace = \real^3 \times \rotmat(3)}$. Each arm $i$, where $i=\{L,\;R\}$, interacts with the same object $\object$. In this context, the absolute skill explains the movement of the object $\object$ in the workspace $\workspace = \Askill$, while the relative skill describes the actions of each end-effector $i$ in $\Rskill$, i.e. with respect to the object's reference frame $\{\object\}$. Note that $\{\object\}$ is the centre of the closed-chain dual-arm system.

        The state of the closed-chain dual-arm system in the workspace can be described by the position/orientation, linear/angular velocities and accelerations of $\{\object\}$ in $\Askill$. As introduced previously, the system's state transition is subjected to its modelled dynamics. \fref{fig:system} illustrates the proposed modelisation of the system's dynamics in $\Askill$ as a set of \acp{DMP} acting between the objects's frame $\{\object\}$ and its goal configuration $\vgoal_o$, which accounts for a desired goal position $\vgoal_{o_x} \in \real^3$ and orientation $\vgoal_{o_q} \in \real^4$. Therefore, three positional \acp{DMP} as in \eref{eq:dmp_1}-\eref{eq:dmp_2} and one orientational \ac{DMP} as in \eref{eq:q_dmp_1}-\eref{eq:q_dmp_2} are required to encode the system's dynamics in the absolute space ${\Askill = \real^3 \times \rotmat(3)}$. 
        
        In the relative space $\Rskill$, the dynamics of each end-effector are modelled as \acp{DMP} referenced to the objects's \mbox{frame $\{\object\}$}. Since ${\Rskill = \real^3 \times \rotmat(3)}$, each end-effector dynamic's in the relative frame is described by three positional \acp{DMP} as in \eref{eq:dmp_1}-\eref{eq:dmp_2} and one orientational \ac{DMP} as in \eref{eq:q_dmp_1}-\eref{eq:q_dmp_2}. 
        
        Any action referenced to the object's frame can be projected to the end-effectors using the grasping geometry $\graspmatrix$ of the manipulated object. This allows computing the required end-effector control commands to achieve a particular absolute task. A detailed explanation of this transformation can be found in~\cite{pairet2018learning}.

	\section{LEARNING-BASED DUAL-ARM MANIPULATION} \label{sec:framework}
\begin{figure}[b!]
        \centering
        \subfigure[]{\includegraphics[width=8.1cm]{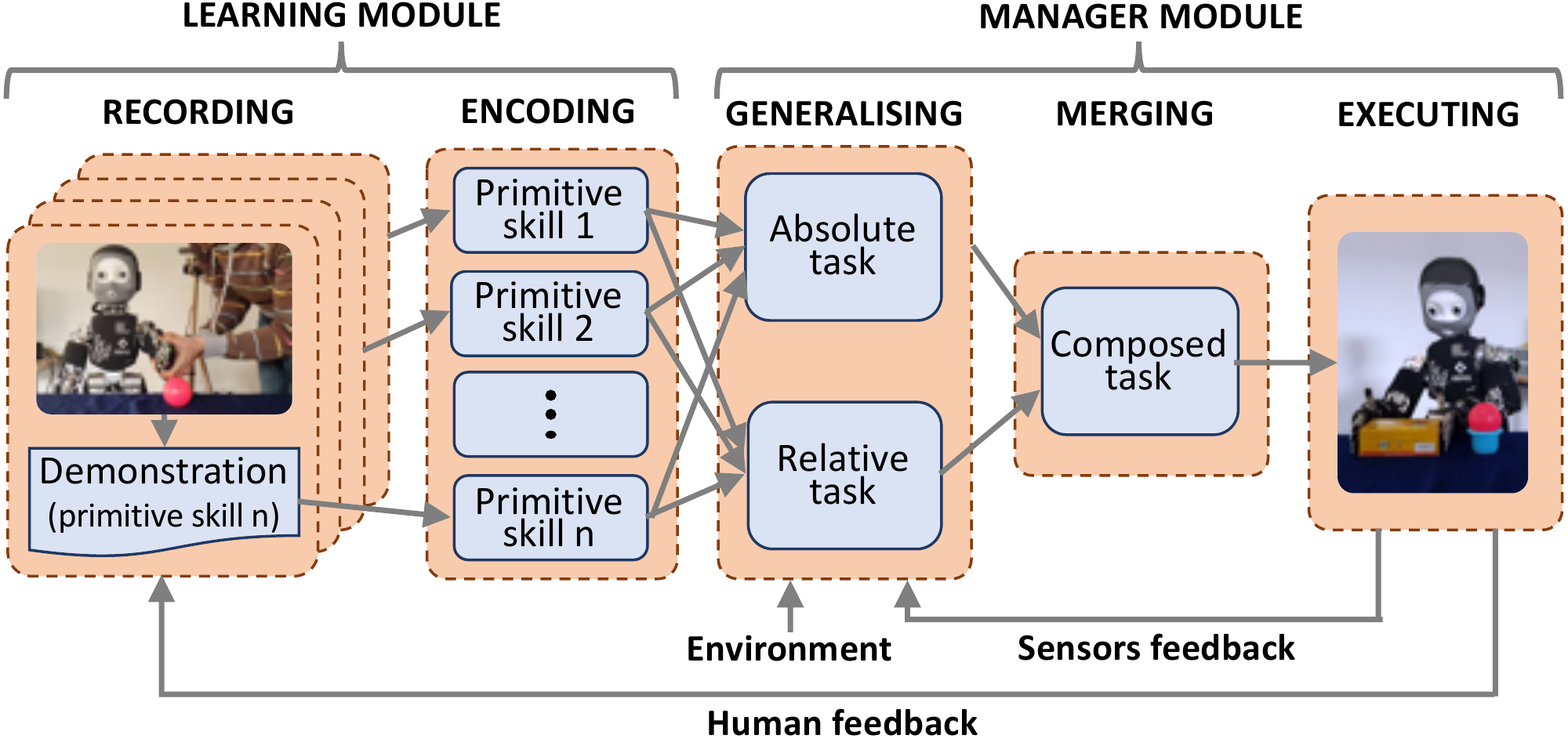}}
        \;
        \subfigure[]{\includegraphics[clip,trim={0cm -1.5cm 0cm 0cm},width=3.5cm]{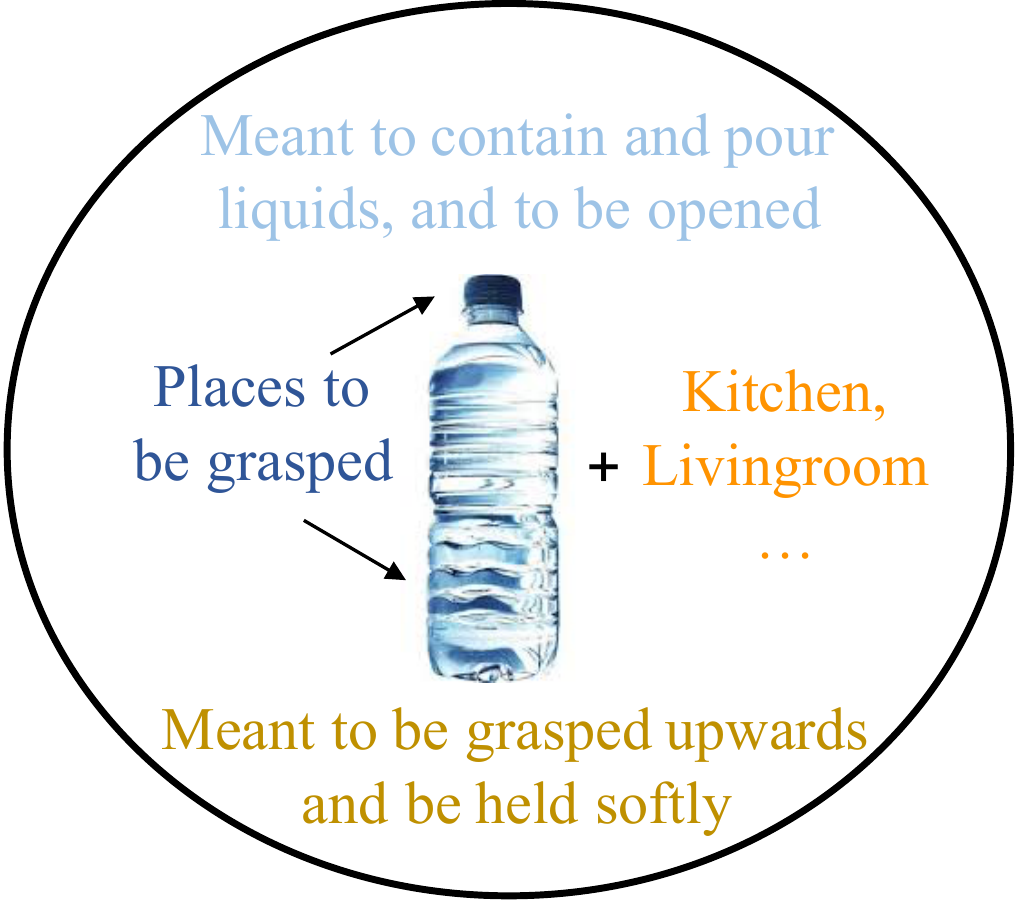} \label{fig:semantics}}
        \caption{Scheme of the proposed framework. (a)~Learning: a library of primitive skills is learnt from human demonstrations. Manager: the primitives are combined simultaneously and sequentially to confront novel environments. (b)~The required primitive skills are selected according to the affordance elements of the dual-arm task.}% of opening a bottle's screw cap.}
        \label{fig:bigpicture}
    \end{figure}

    To endow a dual-arm manipulator with autonomy and robustness in novel scenarios while being easily programmable and customisable by non-robotics-experts, this work has decomposed and modelled the system's dynamics and synchronisation constraints as primitive skills lying in the system's absolute and relative space. Leveraging the formulated modelisation, this work proposes the framework schematised in \fref{fig:bigpicture} which creates and manages a library of primitive skills. The framework has two components: (i)~a learning module that learns a set of primitive skills from human demonstrations, and (ii)~a manager module that combines simultaneously and sequentially these primitives to address a wide range of complex tasks in unfamiliar environments.
    
    % ===============================
    % ===============================
    % ===============================
    \subsection{Library Generation}
        A primitive skill is represented by its coupling term and frame of reference, i.e. either absolute or relative. Learning coupling terms only requires a human demonstrator teaching the characteristic skill. As previously introduced, different coupling terms might be better formulated with different mathematical representations, e.g. a weighted combination of non-linear \acp{RBF} to encode the dynamics of a task, an analytical obstacle avoidance expression, or among others, a force profile to control the environmental interaction.

        The modularity offered by the proposed \ac{DMP}-based formulation and its use in two different spaces tackles the hindrance and ambiguity arising when demonstrating all features of a dual-arm task in an all-at-once fashion. This means that instead of learning a task as a whole, the framework harvests a collection of primitive skills. Creating a repertoire of skills referred to as a library, allows the demonstrator to teach in a one-at-a-time fashion, i.e. to focus on one feature of the demonstration at a time~\cite{bajcsy2018learning}. Moreover, this modular library can be employed for movement recognition purposes, where a demonstrated skill can be compared against the existing ones in the library. If the observed behaviour does not match any existing primitive, it is identified as a new skill and can be added to the framework's library~\cite{ijspeert2002movement}.

    % ===============================
    % ===============================
    % ===============================
    \subsection{Attaching Semantics}
        The framework needs additional information to successfully conduct a dual-arm manipulation task. Let us consider the robotic task of opening a bottle's screw cap, where the system needs to select a proper sequence of primitive skills in order to succeed (see \fref{fig:semantics}). This is first a grasping, where each end-effector holds a different component of the bottle, then a synchronous turning referenced in the system's relative space and finally, a placing and releasing primitives. Therefore, in order to ease this action selection, it is essential to attach a semantic description to each primitive skill.

        Semantic labels bridge the gap between the low-level continuous representation of primitives and the high-level description of actions and their influence on objects. An approach to tackle the object affordances challenge consists in combining features from the object and their surroundings to infer on a suitable grasp-action based on their purpose of use~\cite{ardon2019reasoning,ardon2018HRI}. The combination of such elements builds the relationship between context, actions and effects that provide a cognitive reasoning of an object affordance.
        
        % \begin{figure}[t!]
        %     \centering
        %     \subfigure[]{\includegraphics[width=3.5cm]{./Figures/affordances_map.pdf} \label{}}
        %     \subfigure[]{\includegraphics[width=4.3cm]{./Figures/affordances_example.pdf} \label{}}
        %     \caption{Object affordances as described in \cite{ardon2018HRI}. (a)~Affordances map relating \{\textbf{C}\}ontext, \{\textbf{A}\}ctions and \{\textbf{E}\}ffects. (b)~Affordance elements for the examplified dual-arm task of opening a bottle's screw cap.}
        %     \label{fig:semantics}
        % \end{figure}

    % ===============================
    % ===============================
    % ===============================
    \subsection{Library Management}
        Each coupling term stored in the framework's library represents a particular absolute or relative primitive skill. Reproducing a skill consists in using its coupling term as $\adaptation{_{{x}}}$ or $\adaptation{_{{q}}}$ in \eref{eq:dmp_1}-\eref{eq:q_dmp_2}. This computation retrieves the skill's required accelerations, which can be integrated over time to obtain the skill's velocities $\dot{\vy}_{o}$ for an absolute primitive or $\dot{\vy}_{C_i}$ for the end-effector $i$ relative primitive.
        
        The individual retrieval of primitives already accounts for the inner \ac{DMP} generalisation capabilities, such as different start and goal configurations, as well as obstacle locations. However, these primitive skills need to be combined to generate more complex movements, such as a pick-and-place task of a bottle accounting for the presence of unexpected obstacles (absolute space), while opening the bottle's screw cap considering the exerted force (relative space). The presented framework addresses this prerequisite by simultaneously combining different absolute and relative skills as:
        \begin{align}
            \begin{bmatrix}
                \dot{\vy}_L \\
                \dot{\vy}_R
            \end{bmatrix} =
            \graspmatrix^T \sum_{j=1}^J w_j \; \dot{\vy}_{o_j} + \sum_{k=1}^K w_k
            \begin{bmatrix}
                \dot{\vy}_{C_{L,k}} \\
                \dot{\vy}_{C_{R,k}}
            \end{bmatrix},
            \label{eq:framework_main}
        \end{align}
        %where, considering ${i=\{L,\;R\}}$, ${\dot{\vy}_i \in \real^6}$ describes the linear and angular velocity commands for the $i$ end-effector which satisfies the set of activated primitive skills, ${\graspmatrix \in \real^{6 \times 12}}$ is the global grasp map of the two end-effectors grasp matrices as described in~\eref{eq:grasp_matrix}, and ${\dot{\vy}_{o_j} \in \real^6}$ and ${\dot{\vy}_{C_{i,k}} \in \real^6}$ are the velocities of the ${j \in [1, \; J]}$ absolute and ${k \in [1, \; K]}$ relative primitive skill stored in the library. 
        where ${\dot{\vy}_i \in \real^6}$ describes the linear and angular velocity commands of the ${i=\{L,\;R\}}$ end-effector satisfying the set of activated primitive skills, ${\graspmatrix \in \real^{6 \times 12}}$ is the global grasp map of the two end-effectors grasp matrices as described in~\cite{pairet2018uncertainty}, and ${\dot{\vy}_{o_j} \in \real^6}$ and ${\dot{\vy}_{C_{i,k}} \in \real^6}$ are the velocities of the ${j \in [1, \; J]}$ absolute and ${k \in [1, \; K]}$ relative primitive skill stored in the library. Absolute and relative skill selection is conducted with the weights $w_j$ and $w_k$, respectively.
        
        %Bearing in mind the orthogonality between absolute and relative spaces and the need of avoiding interference between skills laying in the same space,
        
        The resulting framework does not only combines skills simultaneously, but also sequentially. This allows the execution of a complex task composed of a sequence of primitives. To do so, a primitive skill is executed by initialising it with the full state (pose, velocities and accelerations) of its predecessor primitive skill. Such an initialisation avoids abrupt jumps in the system's state.

	\section{RESULTS and EVALUATION} \label{sec:results}
    The proposed framework has been evaluated on the iCub humanoid robot. Particularly, the real platform has been used to load the framework's library with a set of primitive skills learnt from human demonstrations. These skills have been employed in simulation to conduct dual-arm pick-and-place tasks of a parcel in novel scenarios, demonstrating the proposal's potential for humanoid robots. %This section firstly introduces the deployment of the framework on the iCub robot. It then overviews the procedure to teach a robot via human demonstrations. Finally, this section analyses the results of the pick-and-place experiments and demonstrates the potential of the framework for humanoid robots.

    % ===============================
    % ===============================
    % ===============================
    \subsection{Experimental Platform}
        iCub is an open source humanoid robot with 53~\acp{DoF}~\cite{metta2008icub} (see \fref{fig:teaching_b}). The most relevant ones in this work are the three-\acp{DoF} on the torso, the two seven-\acp{DoF} arms equipped with a torque sensor on the shoulder, and the two nine-\acp{DoF} anthropomorphic hands with tactile sensors in the fingertips and palm.
        
        iCub operates under \acs{YARP}. The deployment of the proposed framework on the iCub platform is schematised in \fref{fig:framework}. Mainly, four big functional modules can be distinguished: (i)~the proposed framework described in this paper (blue blocks), (ii)~the real/simulated platform with its visual perception, joint sensors and actuators (magenta blocks), (iii)~the end-effectors control via the built-in \ac{YARP} Cartesian controller~\cite{pattacini2010experimental} and an ad-hoc external torso controller (green blocks), and (iv)~the \acs{HRI} interface to parameterise the desired start and goal configurations for the task, and retrieve the robot's status (red blocks).

    % ===============================
    % ===============================
    % ===============================
    \subsection{Learning Primitive Skills from Demonstration}
\begin{figure}[t!]
            \centering
            \subfigure[]{\includegraphics[width=7.5cm]{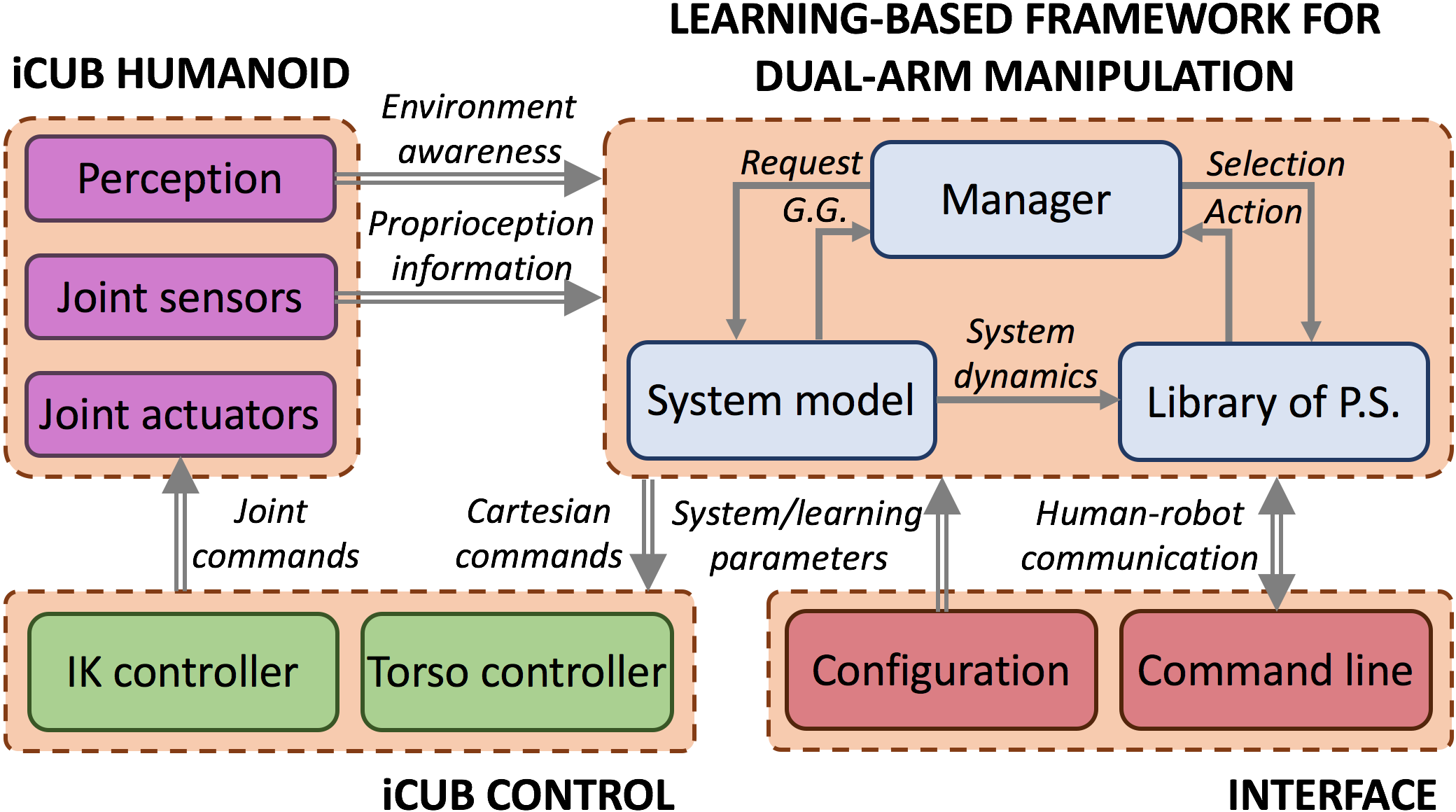}}
            \quad
            \subfigure[]{\includegraphics[clip,trim={0cm -0.5cm 0cm 0cm},height=3.5cm]{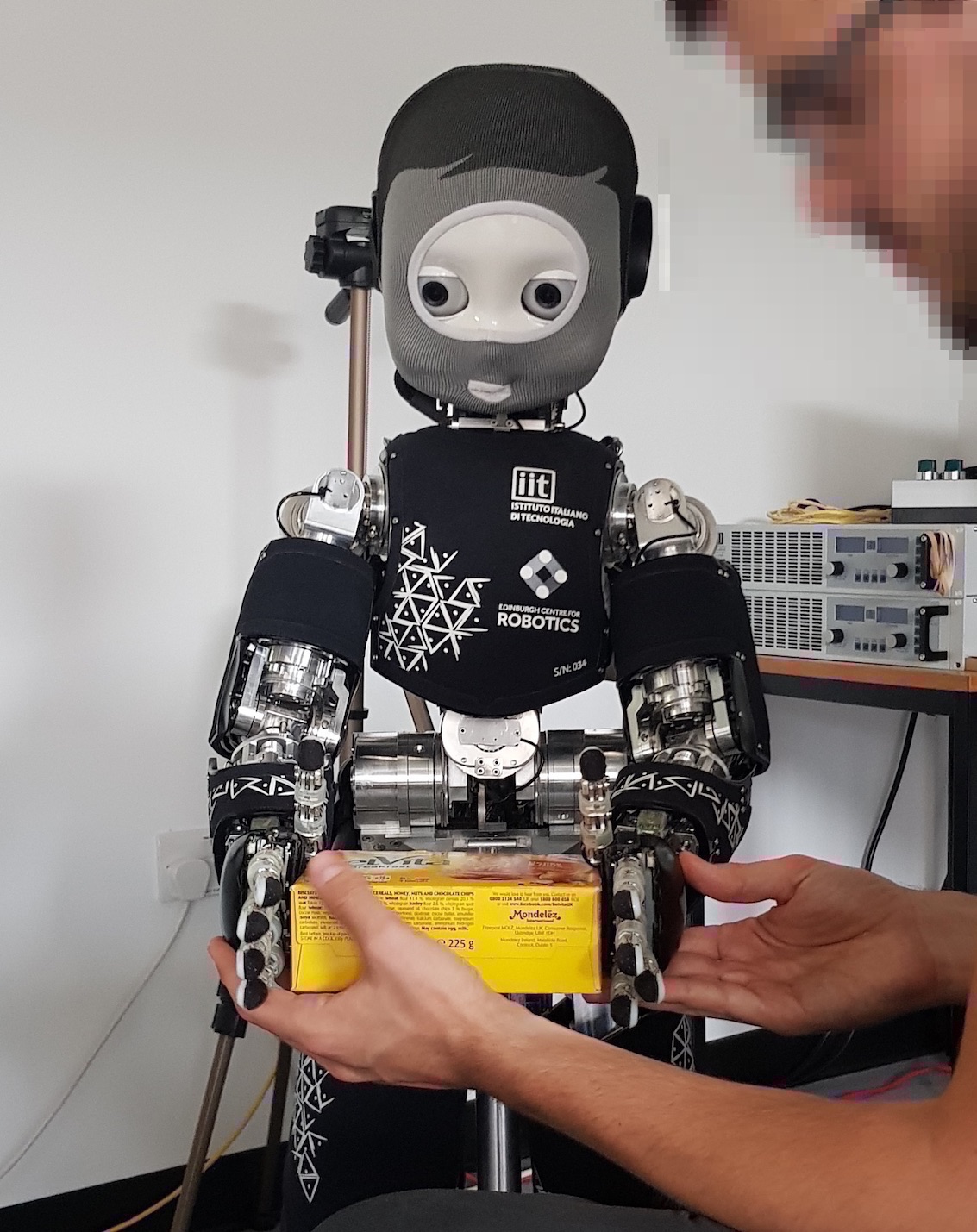} \label{fig:teaching_b}}
            \caption{(a)~Layout of the framework deployment on the iCub robot. Note: grasping geometry (GG), primitive skill (PS). (b)~iCub humanoid being taught grasp maintenance through kineasthetic guiding.}
            \label{fig:framework}
        \end{figure}

        For the system to succeed on the dual-arm pick-and-place of a parcel task in novel environments, the framework's library needs to be loaded with the absolute primitive skills of (i)~pick-and-place dynamics on a horizontal surface, (ii)~rotational motion around the z-axis, and (iii)~obstacle avoidance. Moreover, since the parcel has to be grasped by lateral contact of both end-effectors, the library also requires a relative skill to ensure grasp maintenance, i.e. prevention of contact separation. All these primitive skills have been demonstrated via kineasthetic guiding on the real iCub humanoid robot. To this aim, all joints have been set in gravity compensation, allowing the demonstrator to physically manoeuvre the robot through each primitive. \fref{fig:teaching_b} depicts the kineasthetic teaching of obstacle avoidance and grasp maintenance primitives.
        
        During the demonstrations, proprioception information is retrieved via \acs{YARP} ports to learn the coupling terms $\adaptation{_{{x}}}$ and $\adaptation{_{{q}}}$ in \eref{eq:dmp_1}-\eref{eq:q_dmp_2} characterising the different skills. For the pick-and-place and rotational dynamics, the coupling terms are encoded as a weighted linear combination of non-linear \acp{RBF} distributed along the trajectory as in~\cite{ijspeert2013dynamical}. The obstacle avoidance is learnt by finding the best-fitting parameters of the biologically-inspired formulation as in~\cite{rai2014learning}. Finally, the grasp maintenance skill is learnt by setting the parcel's grasping geometry as a pose tracking reference as in~\cite{gams2014coupling}.

    \subsection{Experiments on Simulated iCub Humanoid}
\begin{figure}[t!]
            \newcommand{\figuresize}{2.4cm}
            \newcommand{\spacesize}{\;}
            \centering
            \subfigure[]{\includegraphics[width=\figuresize]{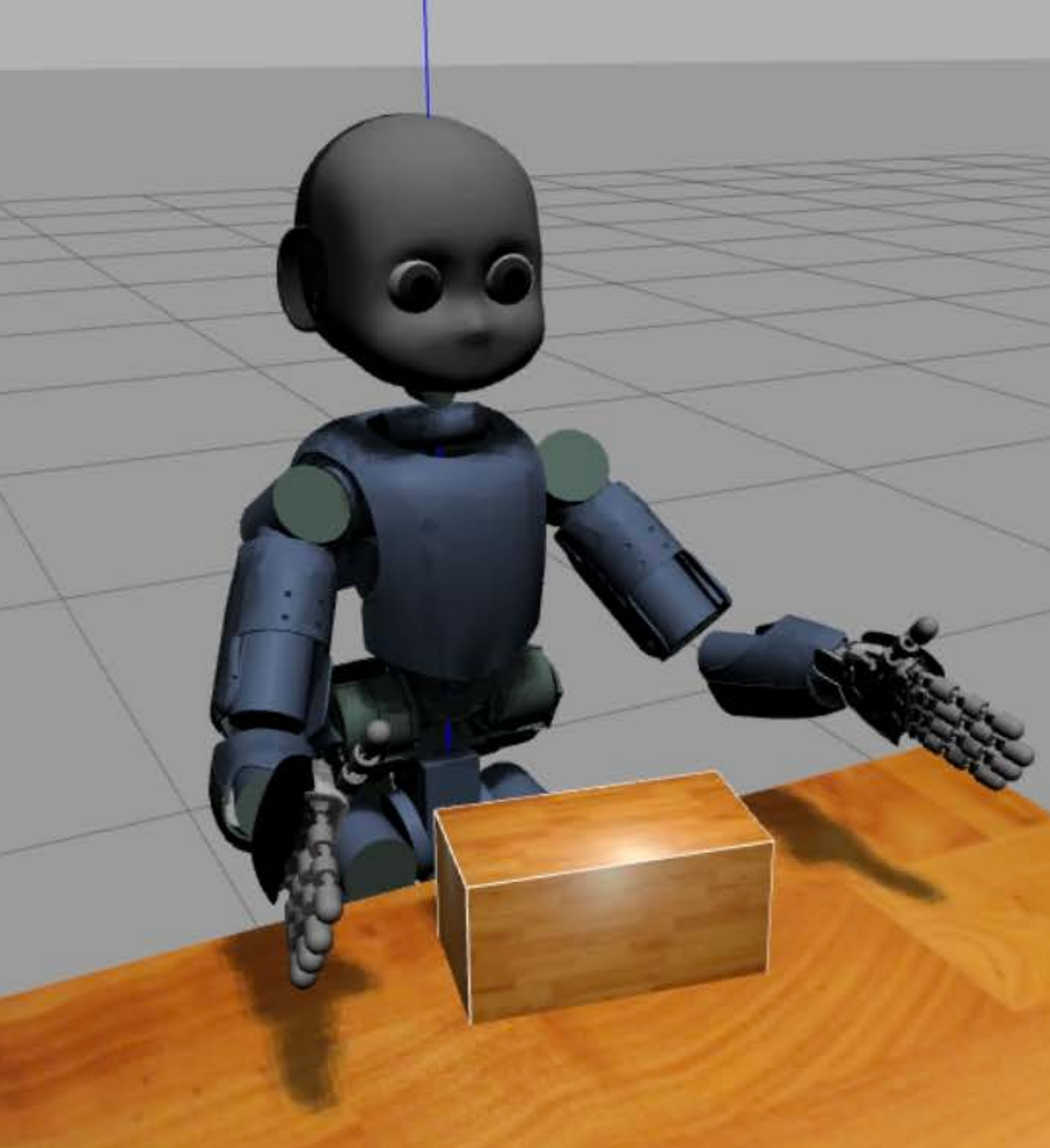}\label{fig:icub_sim_t2_ini}} \spacesize
            \subfigure[]{\includegraphics[width=\figuresize]{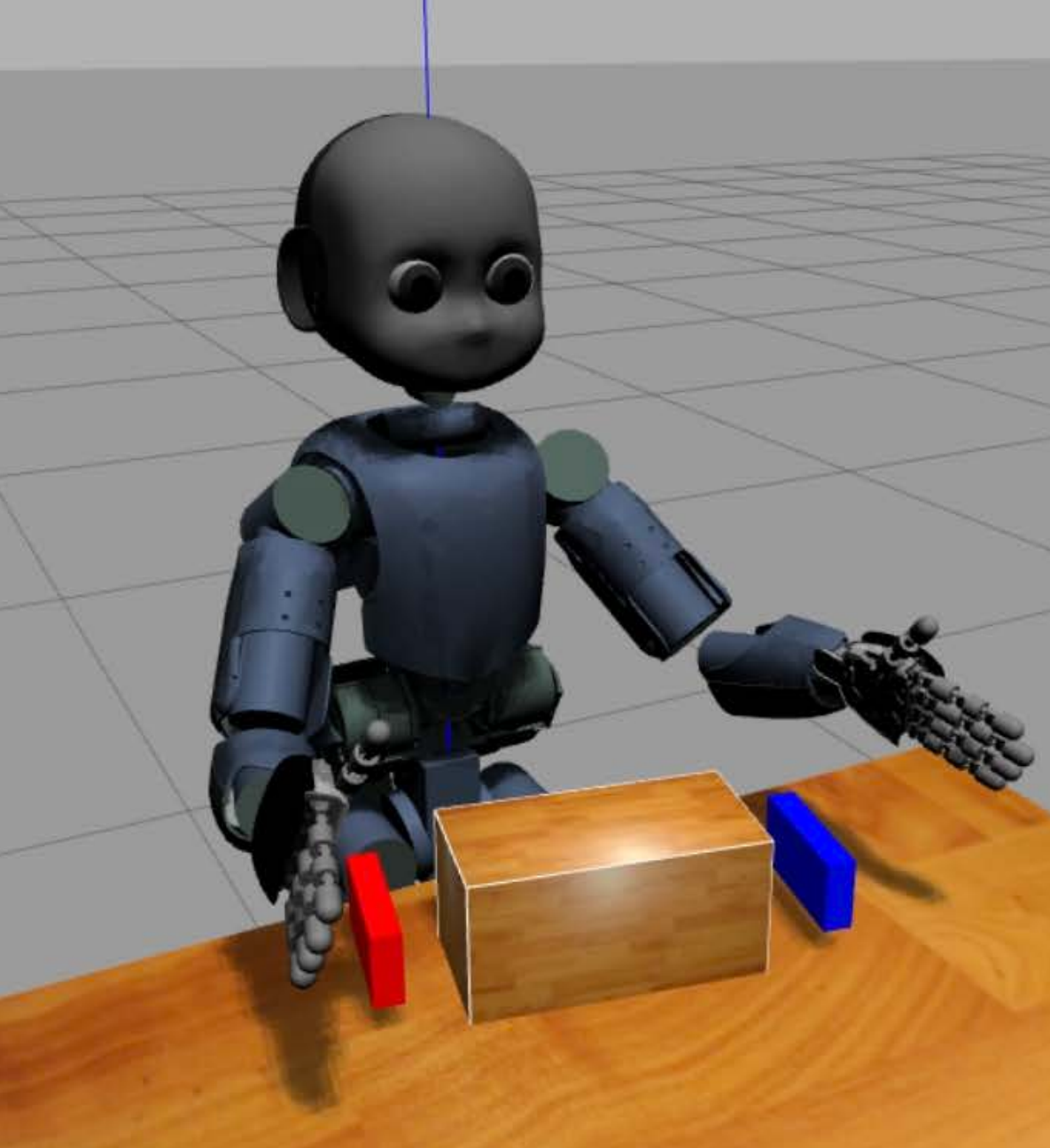}} \spacesize
            \subfigure[]{\includegraphics[width=\figuresize]{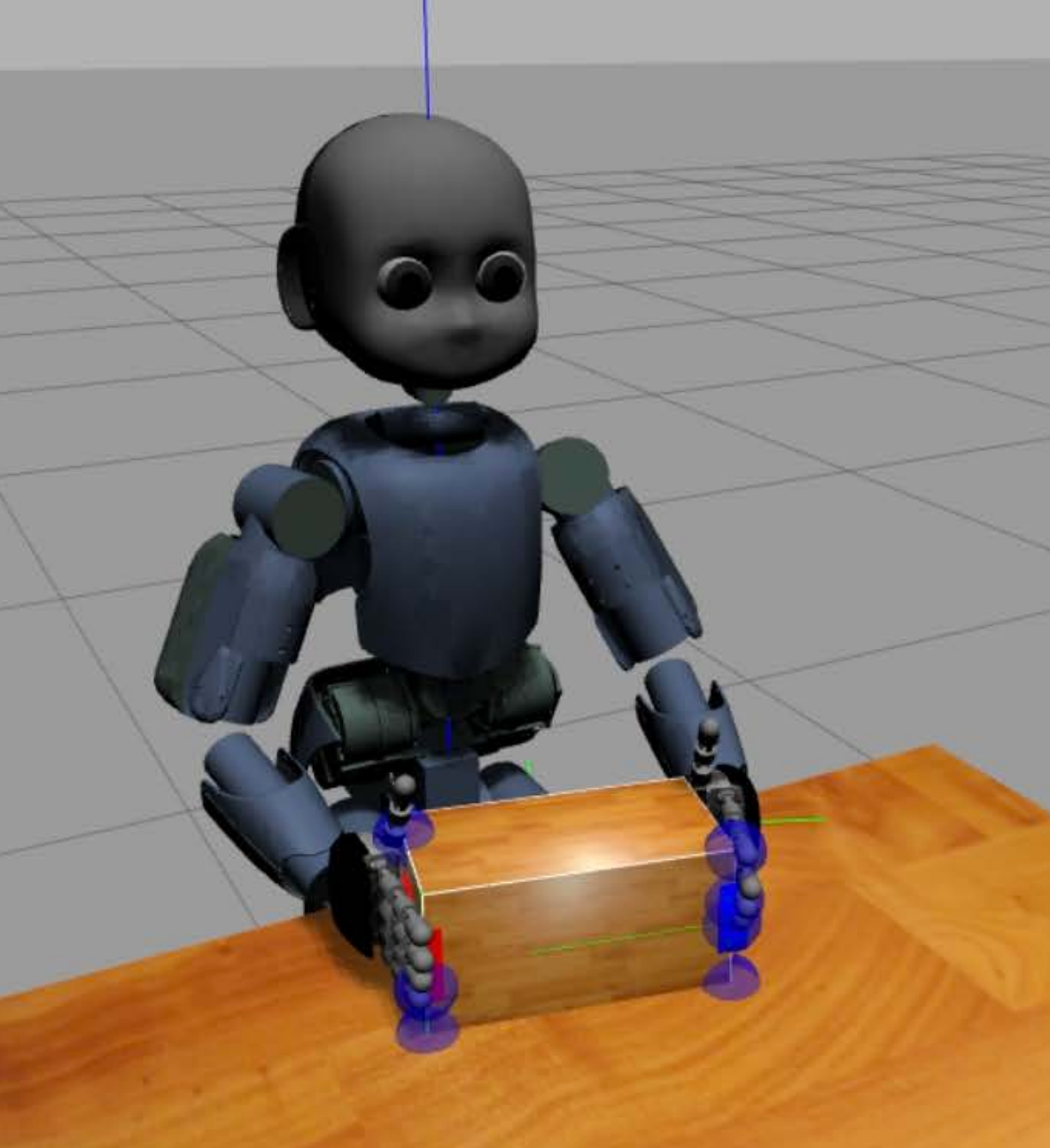}\label{fig:icub_sim_t2_d}} \spacesize
            \subfigure[]{\includegraphics[width=\figuresize]{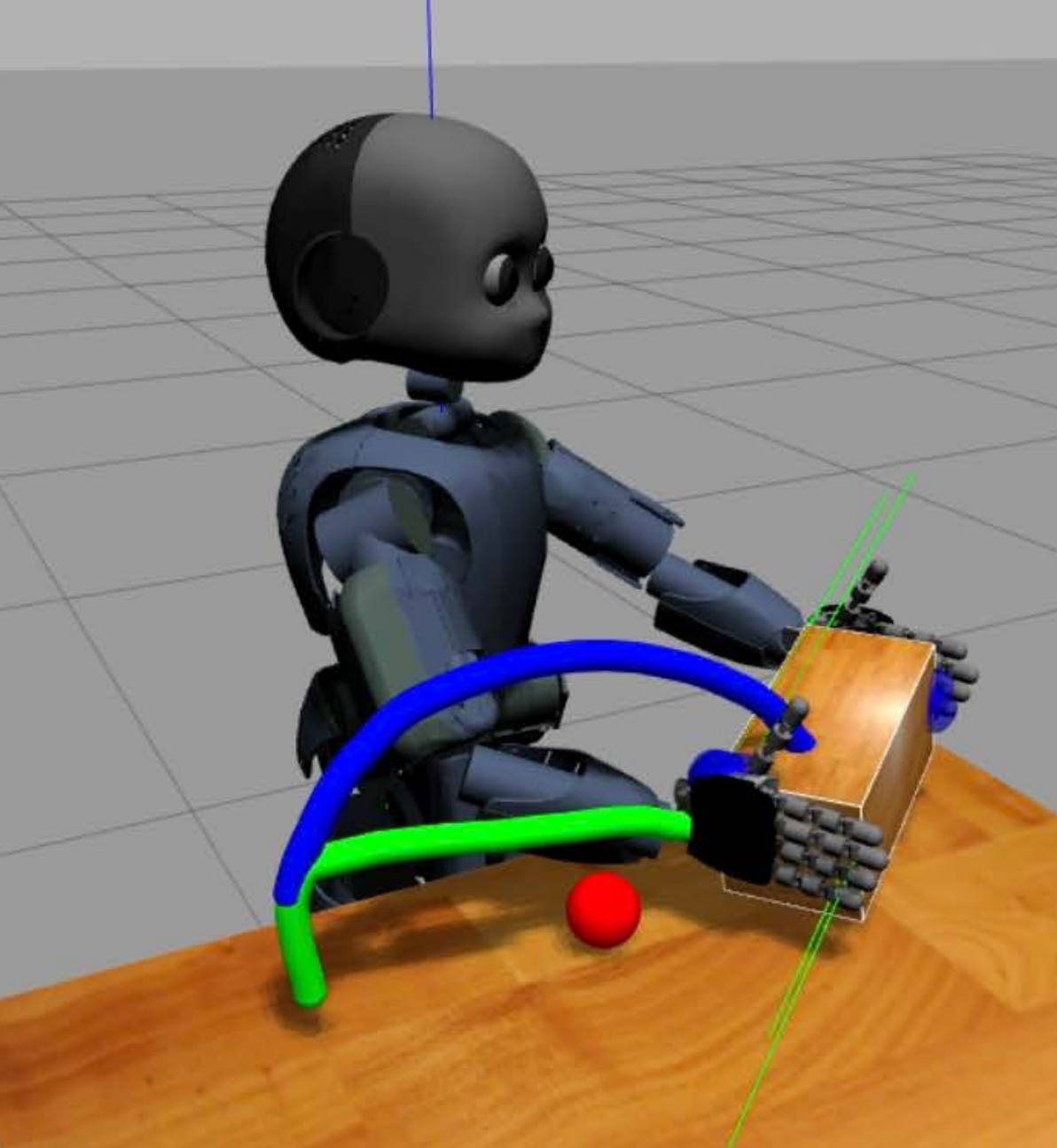}\label{fig:icub_sim_t1_e}} \\
            \subfigure[]{\includegraphics[width=\figuresize]{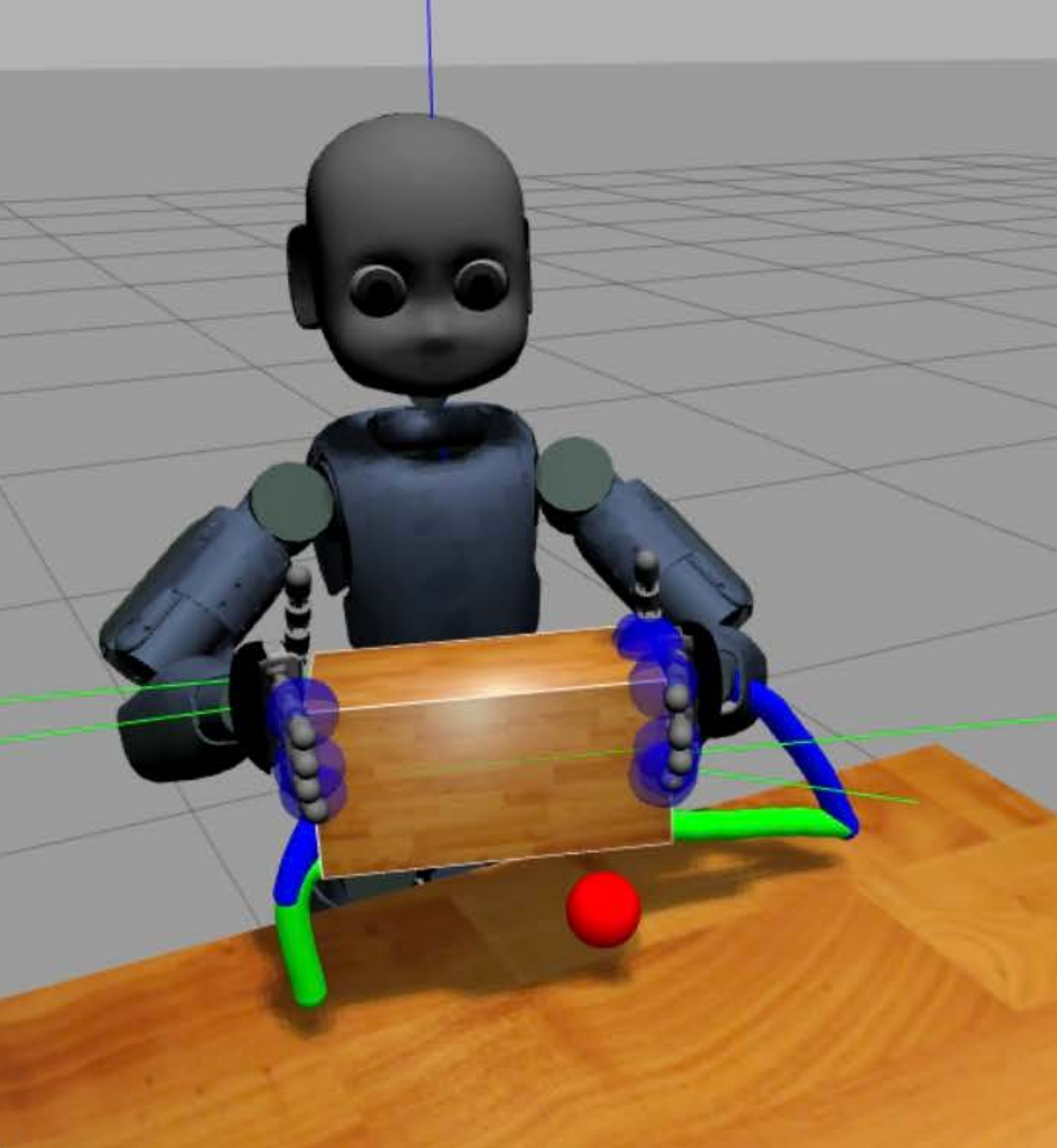}} \spacesize
            \subfigure[]{\includegraphics[width=\figuresize]{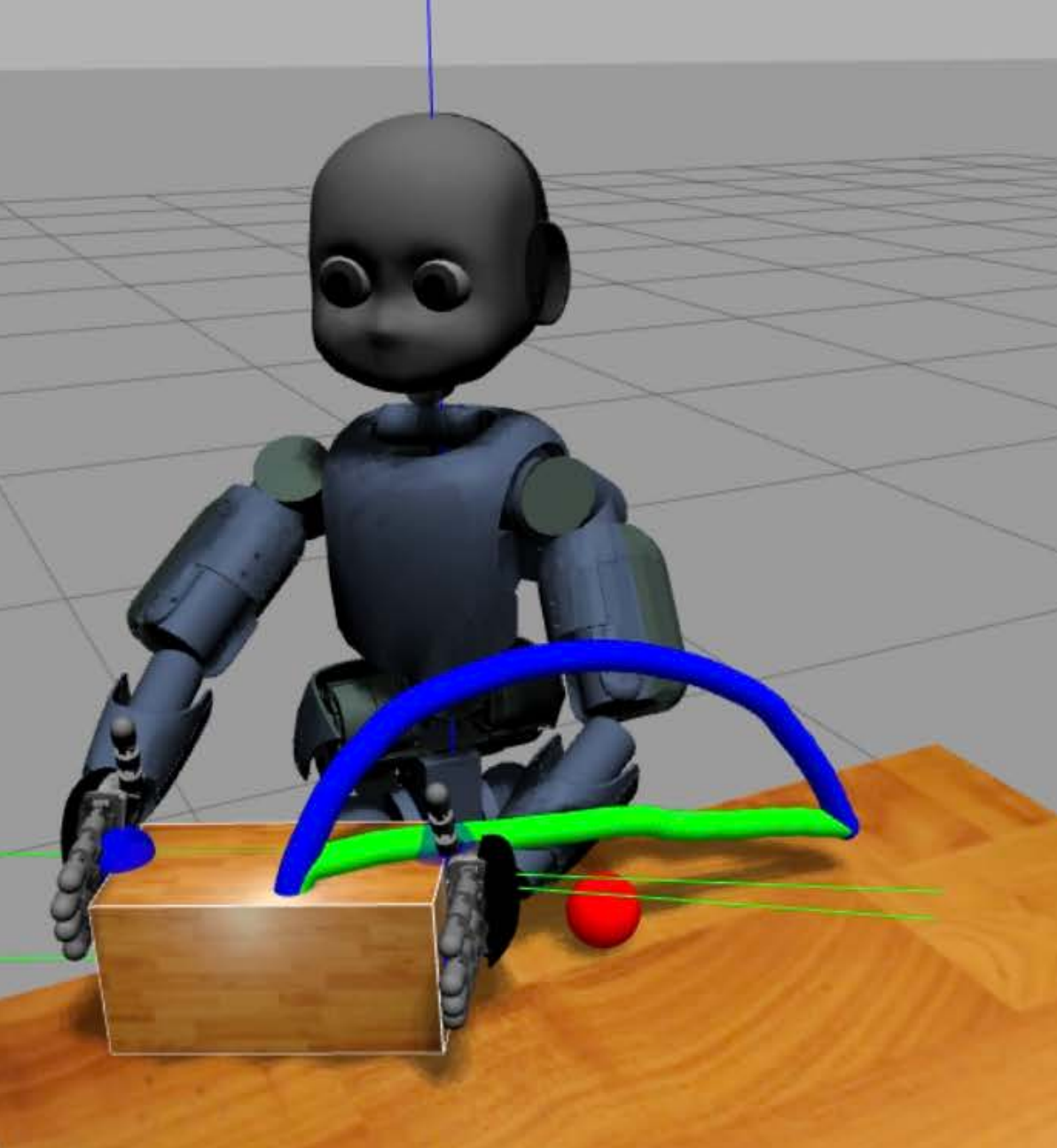}\label{fig:icub_sim_t1_g}} \spacesize
            %\subfigure[]{\includegraphics[width=\figuresize]{Figures/iCub_sim/Binder28.pdf}\label{fig:icub_sim_t2_g}} \spacesize
            \subfigure[]{\includegraphics[width=\figuresize]{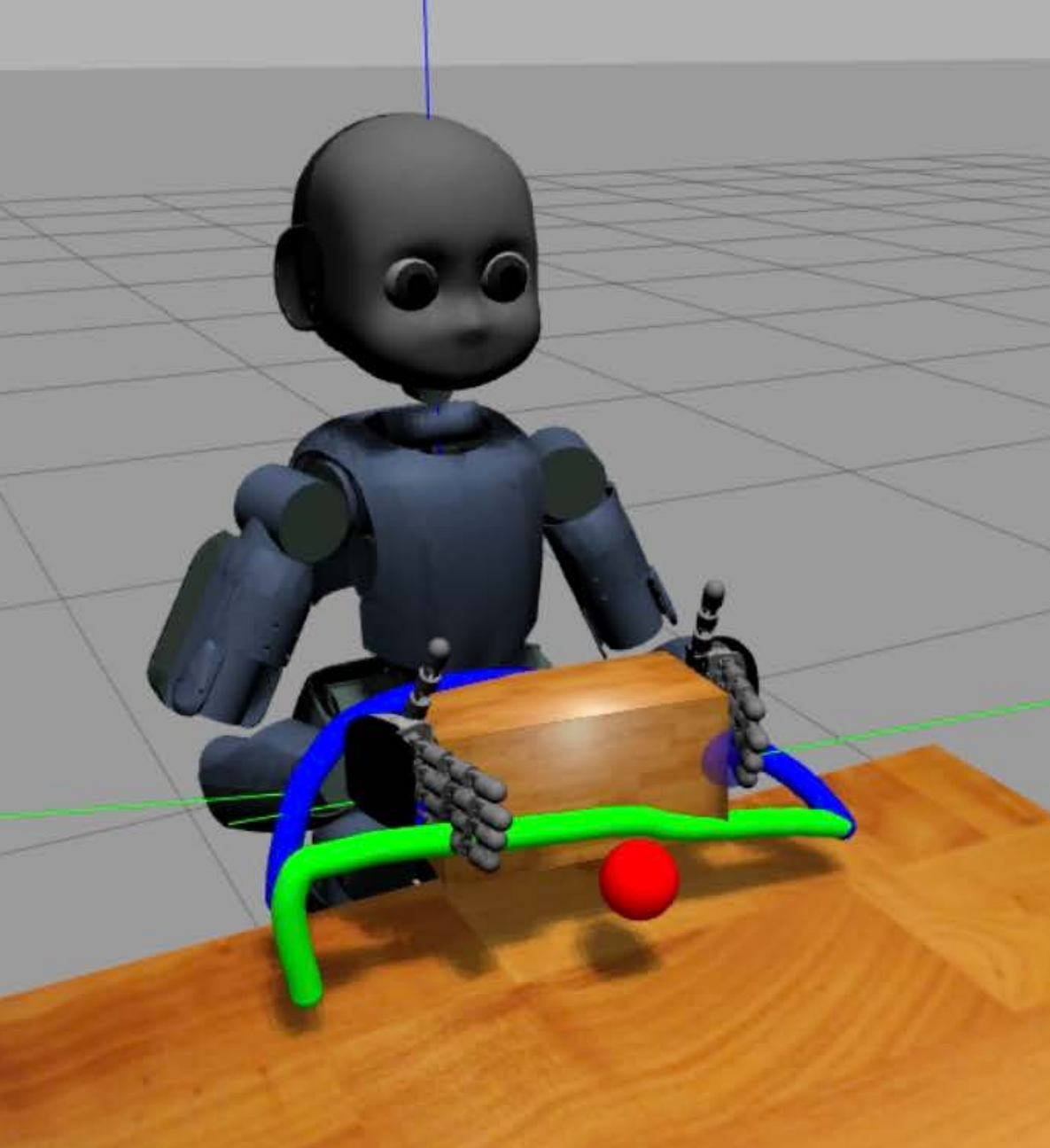}\label{fig:icub_sim_t2_g}} \spacesize
            \subfigure[]{\includegraphics[width=\figuresize]{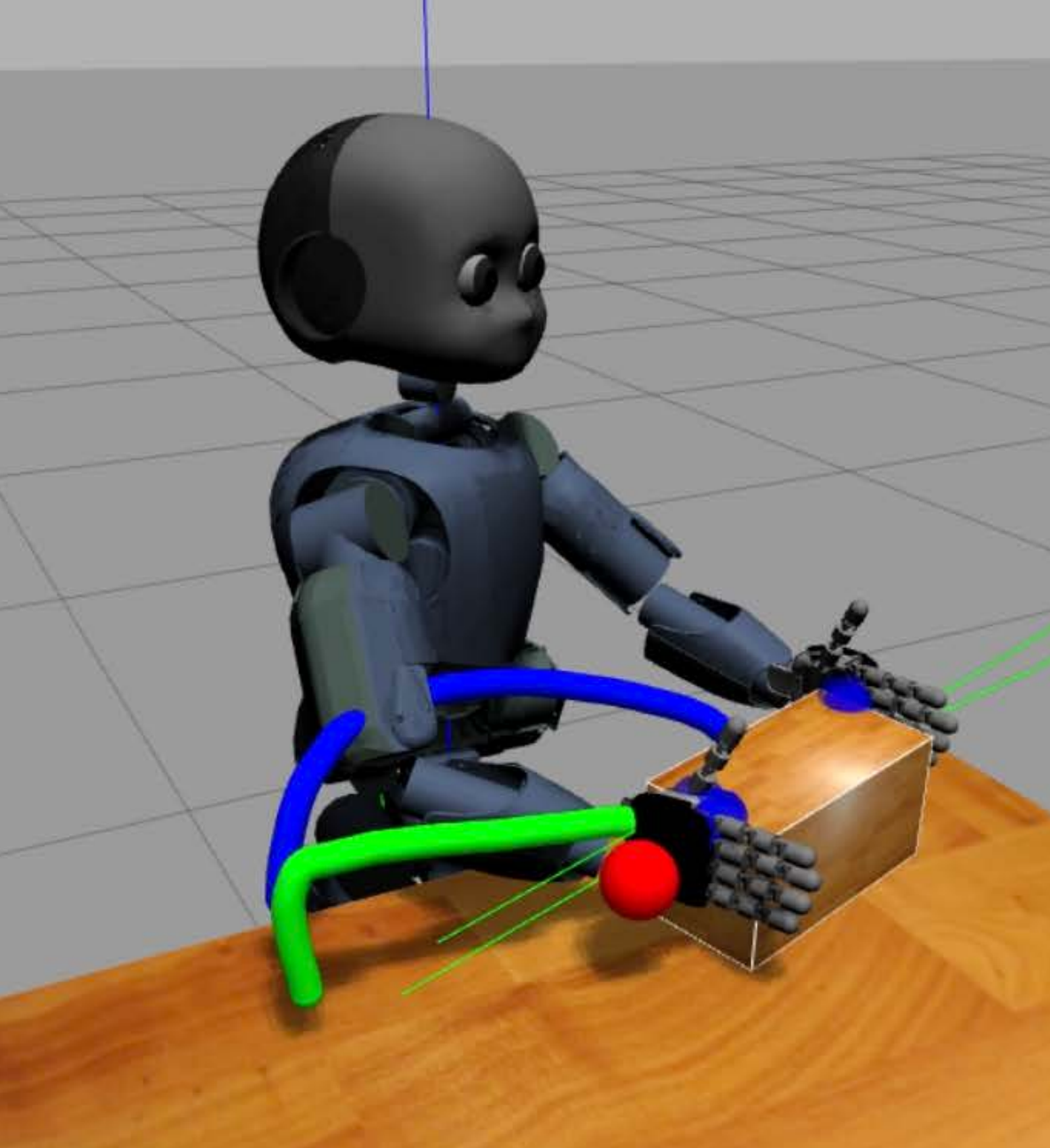}\label{fig:icub_sim_t2_e}}
            \caption{iCub humanoid suceeding in novel dual-arm pick-and-place tasks by simultaneously and sequentially combining primitive skills. Demonstrated pick-and-place (green trajectory). Framework's response (blue trajectory). Obstacle (red sphere). (a)~Parcel initial state. (b)-(c)~Grasping parcel laterally. (d)-(f) and (f)-(h)~Pick-and-place execution with different start, goal and obstacle configurations.}
            \label{fig:icub_sim}
        \end{figure}

        The evaluation of the framework on the pick-and-place setup has been conducted on a simulated iCub robotic platform. Particularly, the four primitive skills previously learnt and loaded in the framework's library are simultaneously and sequentially combined to conduct three consecutive dual-arm pick-and-place task in novel environments (see \fref{fig:icub_sim}).
        
        Given an initial random configuration laying on the table and within iCub's workspace (see \fref{fig:icub_sim_t2_ini}), the first action consists of grasping the parcel. This is achieved by retrieving the parcel's configuration, then use the learnt parcel's geometry to compute the grasping points, and finally approach them laterally via the middle-setpoints displayed as red and blue prisms for the right and left end-effector, respectively (see \fref{fig:icub_sim_t2_ini}-\fref{fig:icub_sim_t2_d}). From this stage on, the grasp maintenance skill ensures that both end-effectors are in flat contact with the box to avoid undesired slippage.

        The following three consecutive movements require picking-and-placing the parcel between different configurations laying on the central, right and left side of iCub's workspace. The former pick-and-place does not require avoiding any obstacle, thus the built-in \acp{DMP} generalisation capabilities are sufficient to address this task (see \fref{fig:icub_sim_t2_d}-\fref{fig:icub_sim_t1_e}). However, the two latter pick-and-place tasks involve adapting the learnt dynamics to address novel scenarios. When the obstacle (red sphere) is collinear with the start and goal positions, i.e. below the demonstrated task (green trajectory), the iCub humanoid circumnavigates the obstacle from the top (see \fref{fig:icub_sim_t1_e}-\fref{fig:icub_sim_t1_g}). Instead, for an obstacle located forward the demonstration, the framework guides the system through a collision-free trajectory near iCub's chest (see \fref{fig:icub_sim_t2_g}-\fref{fig:icub_sim_t2_e}).
        
        The experimental evaluation conducted with the simulated iCub humanoid robot has demonstrated various of the aforementioned framework's features. Having a repertoire of primitive skills available in the framework's library allows exploiting them simultaneously and sequentially to confront complex tasks in novel scenarios. The reported case is one of the 16 successful experiments out of a total of 20 trials. In all cases, the robot had to accomplish the three consecutive dual-arm pick-and-place tasks with different start and goal locations, while avoiding novel obstacles and ensuring grasp maintenance. Failure in any of these tasks made the trial unsuccessful. Interestingly, in the four failed trials one of iCub's forearms collided with the obstacle. This is because the biologically-inspired obstacle avoidance formulation only considers the carried object and should be extended to the object-arm space. The flexibility of the proposed framework could be leveraged to integrate in its library a potential field-inspired approach for obstacle avoidance which also checks for link collisions~\cite{park2008movement}.
	\section{FINAL REMARKS and FUTURE WORK} \label{sec:final_remarks}
    % summary
    This work has presented a novel end-to-end learning-based framework which endows a dual-arm manipulator with real-time and generalisable manipulation capabilities. The framework is built upon the proposed extension of the \ac{DMP}-based modelisation for dual-arm systems, which considers two different frames to reference the movement generation, force interaction and constraints requirements. Based on this arrangement, the proposed framework is twofold: (i)~learns from human demonstrations to create a library of primitive skills, and (ii)~combines such knowledge simultaneously and sequentially to confront novel scenarios. 
    
    % showcase
    The suitability of the proposed approach has been demonstrated in a dual-arm pick-and-place setting, where the iCub humanoid first learnt a repertoire of primitive skills from human demonstrations and then composed such knowledge to successfully generalise to novel scenarios. The framework is not restricted to the presented experimental evaluation nor platform. Any system capable of learning from demonstrations can benefit from this work. Moreover, the framework's modularity allows loading to its library any primitive skill that might be required for dual-arm manipulation purposes.

    % future work
    Future work will significantly extend the library of primitive skills such that more challenging dual-arm manipulation behaviours can be addressed within the framework. In this regard, imminent efforts will focus on learning force-dependant primitive skills or other actions requiring complex synchronisation between end-effectors, such as the opening of a bottle's screw cap or succeeding in the peg-in-a-hole tasks. 
    %Another potential direction for future work is the evaluation of the framework from a \ac{HRI} perspective. Particularly, to assess whether learning composable skills eases the teaching endeavour for naive users, and also increases the system's human-like similarity. Both factors would help in increasing the acceptability and compatibility of robots in human workspaces.

	\section*{ACKNOWLEDGMENTS}
        This work has been partially supported by ORCA Hub EPSRC (EP/R026173/1) and consortium partners.

  \bibliographystyle{Format/splncs04}
  \bibliography{main.bbl}
\end{document}